\documentclass{article}

\PassOptionsToPackage{hidelinks}{hyperref}

\usepackage{PRIMEarxiv}

\usepackage[utf8]{inputenc}
\usepackage[T1]{fontenc}
\usepackage{url}
\usepackage{booktabs}
\usepackage{amsfonts}
\usepackage{nicefrac}
\usepackage{microtype}
\usepackage{lipsum}
\usepackage{fancyhdr}
\usepackage{graphicx}
\usepackage{float}
\usepackage{tikz}
\usepackage{hyperref}

\usetikzlibrary{arrows.meta,positioning}

\AtBeginDocument{\raggedbottom}

\title{How Fast Can Reward Models Score? A Systems Study of C++ and PyTorch Inference Runtimes for RLHF}

\author{
  \textbf{Venkata Naga Sai Vishnu Rohit Pulipaka} \\
  pvnsvrohit@gmail.com\\
  \And
  \textbf{Anish Katta} \\
  anishkatta19@gmail.com\\
  \And
  \textbf{Deva Rohit Reddy Peddireddy} \\
   Devarohitreddy@gmail.com\\
}

\begin{document}
\maketitle

\begin{abstract}
In RLHF pipelines, reward scoring blocks policy updates. Slow scoring bottlenecks the entire loop, since no update runs until every rollout gets a score. And yet most setups just default to PyTorch eager mode or torch.compile, no one checks if that's actually fastest. Scoring itself is small. Rollout generation eats far more of a typical RLHF step. But scoring and generation fight over the same CPU and GPU resources, so a faster scoring engine doesn't shrink step time on its own. It mainly frees up capacity generation can use instead. We built a native C++ inference engine on ONNX Runtime. First step: confirm correctness. Output matched the PyTorch reference to $5.7 \times 10^{-6}$ on CPU and $4.2 \times 10^{-3}$ on GPU, close enough to trust. Then we tested it against PyTorch eager mode, torch.compile, and FastAPI, on both CPU and GPU. CPU was decisive. Our engine beat every baseline, confidence intervals didn't even overlap. GPU gave a different view: we beat PyTorch and FastAPI, but torch.compile came out ahead. Further testing traced the speedup to ONNX Runtime itself, not C++ as a language. And batching strategy mattered more than either the language or the runtime choice, more than we expected. The results are from repeated, independent runs, since single runs just aren't reliable enough to trust.\end{abstract}


\section{Introduction}
  \label{sec:introduction}
  Training models via reinforcement learning from human feedback (RLHF) requires a reward model to score massive batches of candidate responses. This happens at every single training step. Because this inference step is invoked so frequently, per call latency severely impacts total wall clock training time, especially for runs that stretch across multiple days. Surprisingly, reward model serving in most RLHF pipelines is left at framework defaults. Teams typically default to PyTorch eager mode or, occasionally, torch.compile, largely skipping any systematic evaluation of alternative inference backends.

C++ inference engines built on runtimes like ONNX Runtime are standard for production serving outside of the RLHF ecosystem. Yet, there is very little data on whether this approach actually helps here, or by how much, compared to the PyTorch defaults most engineers already use. This paper addresses that gap. We built a custom C++ reward model inference engine, verified its mathematical parity with the PyTorch reference model, and benchmarked its speed against PyTorch eager mode, torch.compile, and a FastAPI serving layer. To ensure these numbers reflect reality rather than a lucky execution environment, every comparison in this study relies on repeated, independent process launches.

Performance diverges sharply depending on the hardware target. On CPU, our ONNX Runtime C++ engine consistently outpaced every PyTorch baseline, with confidence intervals completely detached from the alternatives. The GPU results, however, paint a different picture: while the C++ engine clearly beat plain PyTorch and FastAPI, torch.compile pulled ahead at the median. We confirmed this held up as a statistically significant effect.

Beyond the choice of engine, two deployment level factors proved critical. Padding every batch to a fixed length, rather than grouping similar length requests first, cost 5 to 8 times the throughput on CPU and 3.5 to 4 times on GPU. Length aware grouping recovered that loss only on GPU, since our CPU environment lacked meaningful batch parallelism. Furthermore, funneling multiple requests into a single shared engine instance yielded no throughput gain; execution serialized completely regardless of concurrent arrival. 

Crucially, every metric reported here derives from independent process launches. Run to run noise on a single machine was substantial enough to falsely present one system as faster if left unchecked.
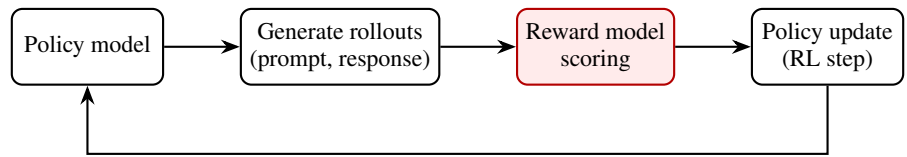
\begin{figure}[H]
\centering
\begin{tikzpicture}[
    node distance=1cm and 1cm,
    box/.style={rectangle, rounded corners, draw, thick, minimum width=2cm, minimum height=1cm, align=center, font=\small},
    highlight/.style={box, draw=red!70!black, thick, fill=red!8},
    arrow/.style={-{Stealth[length=2.5mm]}, thick}
]
\node[box] (policy) {Policy model};
\node[box, right=of policy] (rollout) {Generate rollouts\\(prompt, response)};
\node[highlight, right=of rollout] (reward) {Reward model\\scoring};
\node[box, right=of reward] (update) {Policy update\\(RL step)};

\draw[arrow] (policy) -- (rollout);
\draw[arrow] (rollout) -- (reward);
\draw[arrow] (reward) -- (update);
\draw[arrow] (update.south) -- ++(0,-0.9) -| (policy.south);

\node[below=0.2cm of reward, font=\footnotesize\itshape, red!70!black]{};
\end{tikzpicture}
\caption{The RLHF training loop. Every rollout must be scored by the reward model before the policy update, making its inference latency a direct multiplier on training wall-clock time. This paper studies how fast that scoring step can run.}
\label{fig:rlhf-loop}
\end{figure}

\section{Related Work}
\label{sec:related-work}

RLHF became widely adopted after Christiano et al. \cite{christiano2017deep} demonstrated that human preference comparisons could guide reinforcement learning without a hand coded reward function. Stiennon et al. \cite{stiennon2020learning} expanded this paradigm to summarization, while Bai et al. \cite{bai2022training} applied it to open ended dialogue. In fact, their hh rlhf dataset supplies the prompts and responses scored in our benchmarks. The technique became standard for instruction tuned models following the InstructGPT architecture \cite{ouyang2022training}, which optimizes the policy using proximal policy optimization \cite{schulman2017proximal}.

Much of the subsequent RLHF literature has zeroed in on the reinforcement learning mechanics: sampling strategies, policy training stability, and mitigating reward hacking \cite{gao2023scaling,casper2023open}. How fast the reward model actually executes during these thousands of steps has largely been ignored. Even the existing systems literature focuses overwhelmingly on generation rather than scoring. For example, DeepSpeed Chat \cite{yao2023deepspeedchat} pinpoints actor generation as the dominant cost in an RLHF step, prompting their Hybrid Engine design. Xiao et al. \cite{xiao2023adaptive} similarly observe that generation accounts for over 85 percent of step time, leaving roughly 10 percent for the training update. Neither study isolates reward scoring latency as a distinct bottleneck, which is exactly the gap our work fills (we revisit their timing breakdowns in Section 5).

There is a massive body of literature on inference optimization, but it is aimed elsewhere. Graph based runtimes like ONNX \cite{bai2019onnx} accelerate production inference by shifting models out of Python into dedicated serving environments. PyTorch torch.compile \cite{ansel2024pytorch} achieves similar ends natively by compiling eager mode code into fused kernels. Meanwhile, serving frameworks such as Clipper \cite{crankshaw2017clipper}, Orca \cite{yu2022orca}, and vLLM \cite{kwon2023efficient} emphasize batching and scheduling under heavy concurrent traffic, primarily for generative models bottlenecked by token decoding. None of this prior work asks whether a purpose built C++ engine outperforms PyTorch defaults for the computationally simpler task of scoring a reward model, nor whether generation centric batching behavior applies to a standard forward pass. 

Our findings clarify that the true performance divide is graph execution versus eager mode, rather than C++ versus Python. ONNX Runtime and torch.compile exist specifically to reclaim the overhead left behind by eager PyTorch. Our isolation tests, running the same ONNX session via Python instead of C++, confirm the gains stem entirely from the execution mode.

Finally, our methodology heavily leans on foundational work regarding performance measurement. Mytkowicz et al. \cite{mytkowicz2009producing} and Georges et al. \cite{georges2007statistically} demonstrated that uncontrolled variables, such as memory layout or OS scheduling, can falsely crown a slower system as the winner. Consequently, we report all data as means and confidence intervals across independent process launches, strictly avoiding single run measurements.

\section{Methodology}
\label{sec:methodology}

\subsection{Engine and Models}
Built around ONNX Runtime, our C++ engine takes over tokenization, batching, and postprocessing, jobs that used to live in Python. Two transformer encoder reward models \cite{vaswani2017attention} were tested, both built on the BERT pretraining approach \cite{devlin2019bert}: OpenAssistant's DeBERTa v3 large reward model \cite{he2021debertav3} as the main target, with an Electra large discriminator reward model \cite{clark2020electra} added as a second check, since a single architecture wouldn't tell us whether the findings generalized. The two models also diverge on tokenization (DeBERTa uses SentencePiece \cite{kudo2018sentencepiece}, Electra uses WordPiece \cite{wu2016google}), and the C++ engine implements both natively rather than calling out to Python for either. After export to ONNX from their original PyTorch checkpoints, inference inside the C++ engine runs entirely through the exported graph (no PyTorch involved at that point).

\subsection{Correctness Validation}
Fast and wrong helps no one, so before any speed comparison, we confirmed the C++ engine computed the same thing as the PyTorch reference model. Identical inputs went through both, and reward scores were compared directly. On CPU, the largest absolute difference was 5.7e-6 (mean 3.8e-6); on GPU it was larger, 4.2e-3 at the max and 1.9e-3 on average, a gap expected rather than a bug, since CUDA kernels accumulate floating point operations in a different order than CPU kernels do. Every latency number in later sections only matters because this check passed first.

\subsection{Baselines}
Three PyTorch execution paths serve as control baselines: Hugging Face Transformers eager mode \cite{wolf2020transformers} (representing the default most RLHF codebases reach for without a second thought), torch.compile \cite{ansel2024pytorch} (which fuses operations into optimized kernels without a separate export step), and a FastAPI service wrapping the PyTorch model. That last one stands in for a realistic serving layer, since much of RLHF infrastructure calls the reward model over HTTP rather than in process.

\subsection{Statistical Method}
Because process to process noise on a single machine, OS scheduling quirks, CPU turbo frequency scaling, is large enough by itself to make one system look faster than another when nothing in the code actually changed, a single timed run isn't a reliable measurement. So rather than run each system once and collect internal trials, we launched each as a fresh, independent process multiple times: five repeats for the C++ engine on both CPU and GPU, five for all three PyTorch baselines on GPU. On CPU, those same baselines ran three times each instead of five (torch.compile recompiles for every new input shape, and a full five run CPU sweep would have cost real time against an interval that was already tight). One consequence: the three launch CPU intervals are more sensitive to a single unusual launch than the five launch intervals used elsewhere. This doesn't appear to change the CPU conclusion, the closest CPU baseline's mean sits many standard deviations from the C++ engine's, but these particular intervals should be read as less stable than the rest.

Using Python's \texttt{statistics} module with a fixed lookup table of small sample t critical values (rather than \texttt{scipy}), we computed mean, standard deviation, and confidence intervals throughout. As a supplementary check for this revision, a two sample Welch's t-test (unequal variances, via \texttt{scipy.stats.ttest\_ind}) was also run for each headline comparison, reported alongside the confidence intervals in Section~\ref{sec:main-latency-results}.

Within a single launch, every row in the evaluation set gets scored once, and the median (p50) and 95th percentile (p95) of those per row latencies become that launch's summary values. What Table~\ref{tab:main-latency} and the other latency tables report, then, is not a percentile pooled across all rows from all launches, but the mean and 95 percent confidence interval (small sample t distribution) of the per launch summary value across independent launches: a mean of medians for p50, a mean of 95th percentiles for p95. A result only counts as real if the confidence intervals of the two systems being compared don't overlap.

\subsection{Dataset}
The data is from Anthropic's hh-rlhf dataset, all benchmarks ran over the same prompts and responses. The main evaluation set is 60 rows sampled with a fixed seed, chosen intentionally to cover a range of response lengths. No formal a priori power analysis was run, since that framework doesn't map cleanly onto a repeated independent process launch design; instead we reasoned informally about the scale needed, then checked that reasoning empirically rather than leaving it as an assumption.

In practice, the per launch summary statistics over 60 rows proved tight enough that launch to launch noise stayed small relative to the between system differences reported. The C++ engine's CPU p50, for instance, varied by only about 9 percent of its mean across 5 launches (standard deviation 30.7 ms on a mean of 335.9 ms), against a gap to the nearest CPU baseline of roughly 246 ms (about 8x that noise). To rule out an artifact of which 60 rows happened to get sampled, or of 60 rows simply being too small, two additional 60 row samples with different seeds reproduced the core findings, and an independently sampled 150 row set (roughly 2.5x the original scale) confirmed the same conclusions held.

\subsection{Batching and Concurrency}
At batch sizes of 1, 2, 4, and 8, we compared naive batching (padding every request to the same fixed length) against length bucketed batching (grouping similar length requests first). On CPU, this ran across all four systems, the C++ engine, HF eager mode, Python ONNX Runtime, and torch.compile, precisely to check the finding wasn't an artifact of one implementation. GPU testing covered the C++ engine only. The full sweep was repeated on Electra, and again on an independently sampled 150 row dataset, as a check against model choice or dataset size driving the result.

For concurrency, requests went out at levels of 2, 4, and 8 against the C++ engine: once with everything sharing a single engine instance, once with each thread given its own (the question being whether a shared instance becomes a bottleneck under load). This ran on both DeBERTa and Electra, on CPU and GPU, plus, for the shared instance case, over real HTTP against a running FastAPI server. As with the main latency comparison, these figures are means with 95 percent confidence intervals across 5 independent launches per configuration. Pinning down the GPU multi instance memory ceiling more precisely required a separate set of single, non repeated checks at concurrency levels 5, 6, and 7, since the main sweep's levels (2, 4, 8) never landed directly on the boundary where independent sessions exhaust GPU memory.

\subsection{Hardware and Limitations}
Every benchmark ran on one development machine, one CPU, one NVIDIA GPU. No second machine with different hardware was available to check generalization, and we'd rather state that plainly than imply a broader hardware claim we haven't tested.

\subsection{Software Versions and Precision}
\label{sec:versions}
Every benchmark, CPU and GPU alike, ran in fp32; no half precision or mixed precision path exists anywhere in the codebase. Both the C++ engine and the Python ONNX Runtime baseline used ONNX Runtime 1.26.0 (CPUExecutionProvider on CPU, CUDAExecutionProvider on GPU), pinned identically across the C++ side (via CMake \texttt{FetchContent}) and the Python environment (via \texttt{pyproject.toml}) so that C++ versus Python parity checks wouldn't be confounded by differing ONNX Runtime releases. PyTorch baselines ran on PyTorch 2.10.0 (the \texttt{+cu126} build for GPU runs), with torch.compile called under no explicit \texttt{mode} argument, meaning the default \texttt{inductor} backend. GPU runs used CUDA 12.6 and driver version 592.00. The CPU is an AMD Ryzen 7 5800H; the GPU is an NVIDIA GeForce RTX 3060 Laptop GPU with 6144 MiB of VRAM (the same card whose memory ceiling resurfaces in Section~\ref{sec:concurrency}). These details matter because both torch.compile's behavior and the ONNX Runtime execution path are sensitive to version and configuration, and reproducing Table~\ref{tab:main-latency} depends on knowing them.

\section{Results}

\subsection{Main Latency Comparison}
\label{sec:main-latency-results}

Table~\ref{tab:main-latency} reports p50 latency on the primary 60 row evaluation set, mean and 95\% confidence interval across independent process launches, all four systems, both devices.

\begin{table}[H]
\centering
\caption{Main latency comparison, p50 (ms), mean [95\% CI] across independent process launches.}
\label{tab:main-latency}
\begin{tabular}{lcc}
\toprule
System & CPU (ms) & GPU (ms) \\
\midrule
C++ engine & 335.9 [297.7, 374.0] & 27.4 [24.3, 30.5] \\
HF eager (PyTorch) & 602.4 [551.3, 653.4] & 57.2 [50.0, 64.4] \\
FastAPI & 581.6 [570.4, 592.9] & 62.8 [52.8, 72.9] \\
torch.compile & 628.8 [613.7, 644.0] & 19.0 [17.6, 20.3] \\
\bottomrule
\end{tabular}
\end{table}

The CPU margin lands between 1.7 and 1.9x at point estimates (335.9 ms for the C++ engine against 581.6–628.8 ms for the three baselines), and even the most conservative single comparison, the C++ engine's upper 95\% CI bound against a baseline's lower bound that still shows at least a 1.4x margin. The C++ engine's confidence interval does not touch any of the three PyTorch baselines. Nothing borderline about it.

GPU execution splits the narrative in two. Plain PyTorch eager mode and FastAPI both lose clearly, with zero interval overlap. However, torch.compile comes in lower at the median, 19.0 milliseconds against 27.4, and the gap holds under the same strict non overlap standard. If you are willing to pay for an initial compile step and tolerate recompilation when input shapes shift, plain PyTorch actually beats a dedicated C++ ONNX Runtime engine on GPU. Stating this plainly seemed far better than trying to explain it away.

\begin{figure}[H]
\centering
\includegraphics[width=0.8\linewidth]{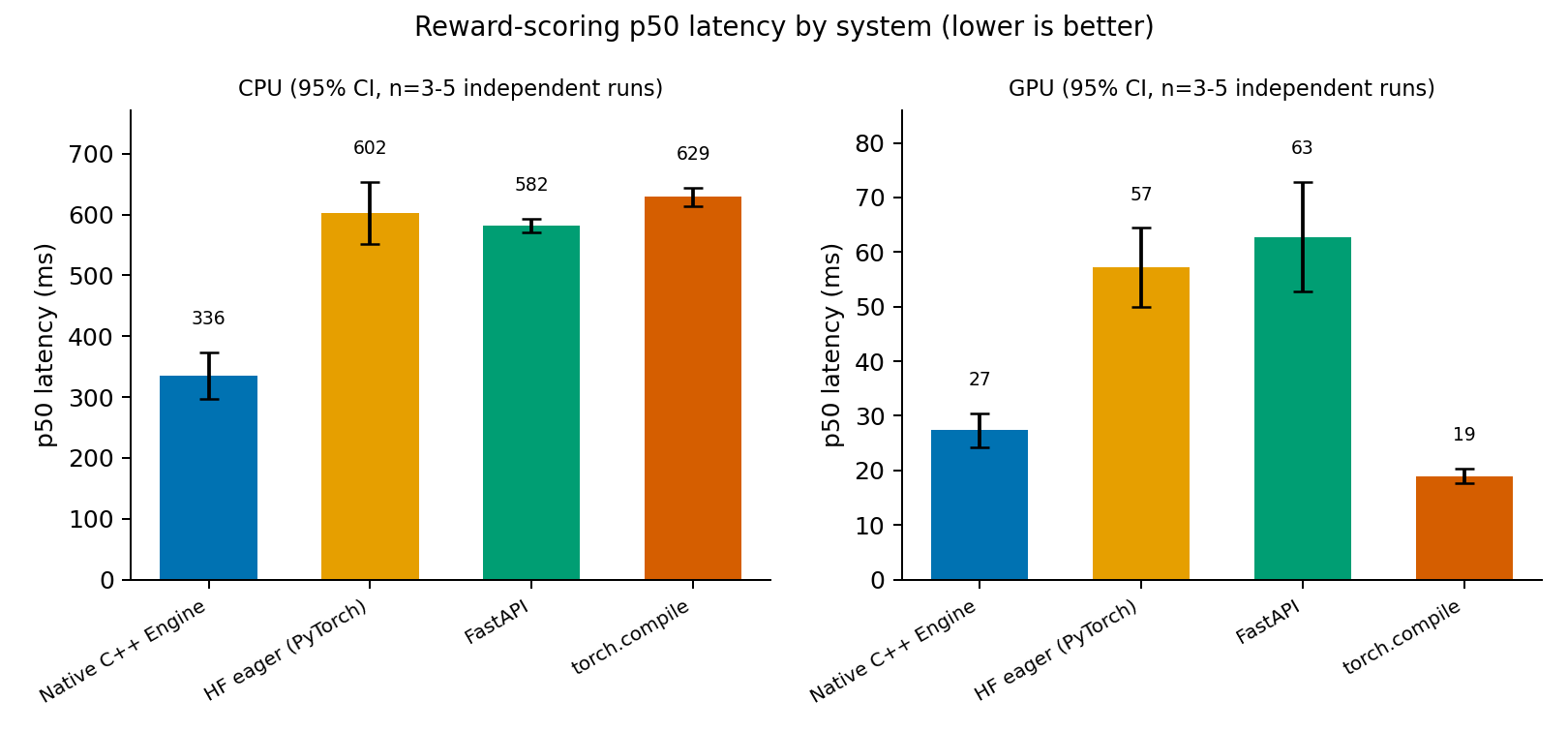}
\caption{p50 latency by system, CPU and GPU, 95\% CI across independent runs.}
\label{fig:main-latency}
\end{figure}

Figure~\ref{fig:main-latency} shows the same comparison as a chart.

A system that wins on average could still lose during a training loop worst case step. To account for this, we checked the tail latencies. Table~\ref{tab:gpu-p95} provides the p95 latency for the same four GPU systems, calculated using the 95th percentile within each launch, followed by the mean and confidence interval across 5 independent launches.

\begin{table}[H]
\centering
\caption{GPU p95 latency (ms), mean [95\% CI] across independent process launches.}
\label{tab:gpu-p95}
\begin{tabular}{lc}
\toprule
System & GPU p95 (ms) \\
\midrule
C++ engine & 116.2 [109.7, 122.8] \\
HF eager (PyTorch) & 148.2 [133.0, 163.5] \\
FastAPI & 140.8 [119.3, 162.3] \\
torch.compile & 25.6 [18.4, 32.9] \\
\bottomrule
\end{tabular}
\end{table}

At the 95th percentile, torch.compile logged 25.6 milliseconds versus the C++ engine 116.2. The gap actually widens at the tail, firmly shutting down the theory that a static ONNX Runtime graph acts as a safer guardrail against worst case latency spikes. On this specific hardware, with this specific workload, it does not. 

Every comparison in this section also clears p < .001 under a two sample Welch t test, which we report in Table~\ref{tab:welch} alongside the confidence intervals to reinforce the non overlap results.

\begin{table}[!htbp]
\centering
\caption{Welch's t-test (unequal variances) for headline C++-engine-versus-baseline comparisons.}
\label{tab:welch}
\begin{tabular}{lccc}
\toprule
Comparison & $t$ & $df$ & $p$ \\
\midrule
C++ engine vs. HF eager (CPU, p50) & $-14.67$ & $5.77$ & $< .001$ \\
C++ engine vs. torch.compile (CPU, p50) & $-20.64$ & $4.50$ & $< .001$ \\
C++ engine vs. FastAPI (CPU, p50) & $-17.56$ & $4.28$ & $< .001$ \\
C++ engine vs. HF eager (GPU, p50) & $-10.53$ & $5.44$ & $< .001$ \\
C++ engine vs. torch.compile (GPU, p50) & $6.87$ & $5.45$ & $< .001$ \\
C++ engine vs. FastAPI (GPU, p50) & $-9.35$ & $4.76$ & $< .001$ \\
C++ engine vs. torch.compile (GPU, p95) & $25.73$ & $7.93$ & $< .001$ \\
\bottomrule
\end{tabular}
\end{table}

\subsection{Why the C++ Engine Is Faster}

Wrong. That's the answer to whether C++ itself explains the CPU advantage, which is the natural assumption and the one we started with. We took the exact same ONNX Runtime session, same model, same library, called it from a plain Python script instead, timed it the same way. About 349 milliseconds, against the C++ engine's 335.9. Look at the confidence intervals rather than the raw numbers and those two are essentially tied, both beating plain PyTorch eager mode's 602.4 milliseconds by a wide margin. Calling ONNX Runtime from Python costs about the same as calling it from C++. The runtime does the work, not the language.

\begin{figure}[H]
\centering
\includegraphics[width=0.4\linewidth]{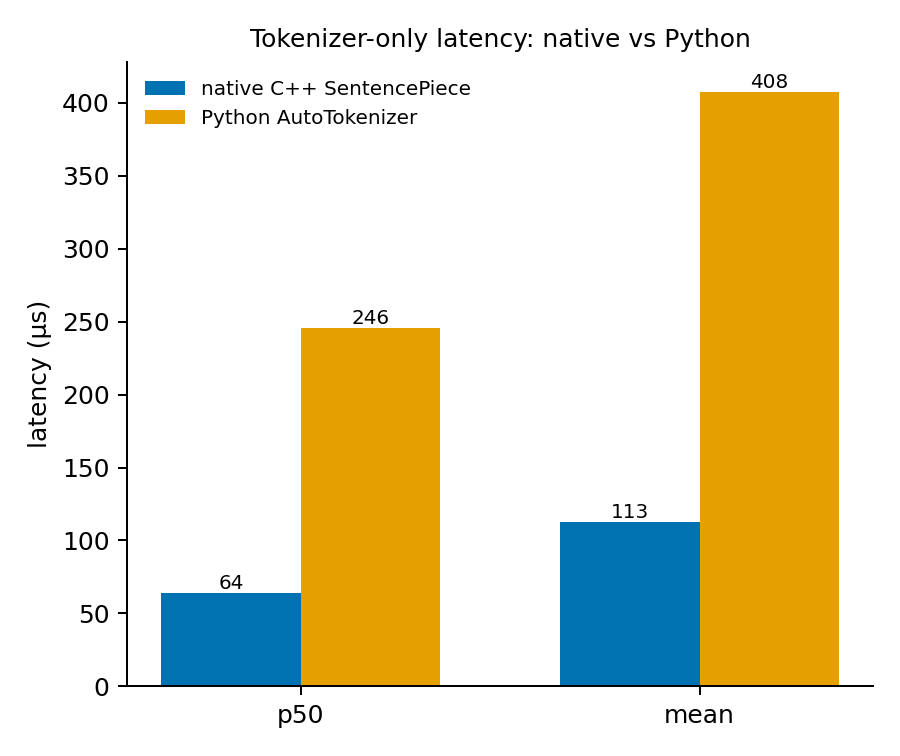}
\caption{Tokenizer-only latency, native C++ SentencePiece vs. Python AutoTokenizer.}
\label{fig:tokenizer}
\end{figure}

So what does C++ buy, if not raw language speed? A small set of ablations answers this. Swapping the native C++ tokenizer for Python's AutoTokenizer produced a real gap, 64.3 microseconds against 245.8, about 3.8 times faster. Small next to a full transformer forward pass, but genuine and repeatable.

Two changes we expected to matter didn't. Building input tensors without an extra memory copy, and preallocating buffers instead of allocating fresh on each call, both showed no effect once checked carefully. An early run of the zero copy change looked like a 10 percent win. It wasn't; the same spread showed up on reruns of the identical baseline with no change applied at all. A roughly one kilobyte copy and one or two heap allocations sit three to five orders of magnitude below a transformer forward pass, so this null result mostly confirms what should have been obvious from the start.

Electra's native tokenizer path told the same story on rerun. Baseline, zero copy, and preallocate all landed in the same range, 228.7 to 238.6 milliseconds on CPU and 16.5 to 18.6 on GPU. Nothing separable from ordinary run to run noise.

\subsection{Batching: Naive Padding vs. Length-Aware Scheduling}

5 to 8x. That's the CPU throughput cost of naive batching, padding every request to the length of its longest batch member, relative to batch size 1 (no batching at all). Table~\ref{tab:batching-cpu} spans batch sizes 1, 2, 4, and 8 across three systems, checking the result wasn't specific to one implementation. Table~\ref{tab:batching-gpu} gives the same comparison on GPU, C++ engine only, the sole system tested there.

\begin{table}[H]
\centering
\caption{CPU batching throughput (rows/sec), naive vs. length-bucketed.}
\label{tab:batching-cpu}
\begin{tabular}{lcccccc}
\toprule
Batch size & C++ naive & C++ bucketed & HF naive & HF bucketed & Py-ONNX naive & Py-ONNX bucketed \\
\midrule
1 & 3.10 & -- & 1.22 & -- & 2.84 & -- \\
2 & 0.61 & 2.89 & 0.43 & 1.86 & 0.53 & 2.19 \\
4 & 0.48 & 2.83 & 0.39 & 2.06 & 0.49 & 2.82 \\
8 & 0.40 & 2.43 & 0.38 & 1.95 & 0.43 & 2.77 \\
\bottomrule
\end{tabular}
\end{table}

\begin{table}[H]
\centering
\caption{GPU batching throughput (rows/sec), C++ engine, naive vs. length-bucketed.}
\label{tab:batching-gpu}
\begin{tabular}{lcc}
\toprule
Batch size & naive & bucketed \\
\midrule
1 & 39.8 & -- \\
2 & 11.1 & 51.5 \\
4 & 10.8 & 52.1 \\
8 & 10.0 & 53.7 \\
\bottomrule
\end{tabular}
\end{table}

Before trusting any of this, we confirmed padding doesn't touch the reward score itself. Batch=1 versus batch=8 was bit-identical on CPU. On GPU the max absolute difference was 0.0016, the same fp32 rounding-order noise already seen between CPU and GPU single-row runs, not a padding bug.

Naive batching actively hurts, it isn't merely neutral, and it gets worse as batch size grows, since a longer batch gives a long row more chances to land in it and drag the whole batch's padding along. Length aware scheduling recovers that loss only on GPU. Bucketed CPU throughput never exceeds the batch=1 baseline of 3.10 rows per second; it plateaus around 2.4 to 2.9 regardless of batch size, because ONNX Runtime's CPU backend folds the batch dimension into one larger sequential matmul rather than running it in parallel, so total compute still scales with total tokens processed. GPU tells the opposite story: bucketed throughput improves roughly 35 percent over batch=1 and keeps climbing with batch size, since spare parallel compute on the GPU lets same length rows actually share work.

The pattern held on rerun. Electra's native C++ path and an independently sampled 150 row dataset (roughly two and a half times the primary sample) both reproduced it: naive batching backfires on both devices, bucketing pays off only on GPU, CPU bucketed throughput never clears its own batch=1 number. The 150 row numbers sat within noise of the 60 row numbers, CPU naive at batch=8 came in at 0.42 rows per second versus 0.40, GPU bucketed at batch=8 came in at 56.4 versus 53.7. Not an artifact of sample size.

\begin{figure}[H]
\centering
\includegraphics[width=0.9\linewidth]{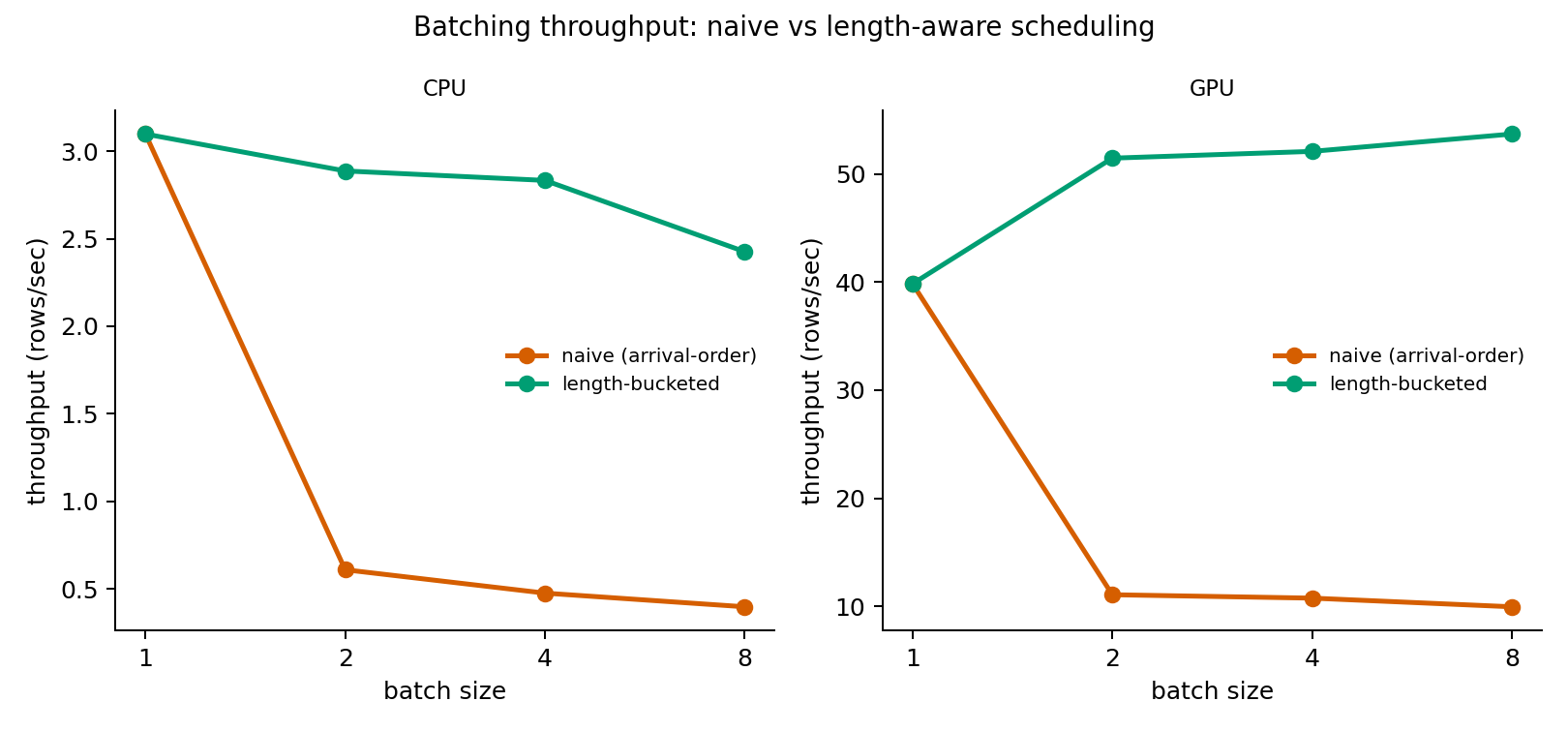}
\caption{Throughput vs. batch size, naive vs. length-bucketed scheduling, C++ engine.}
\label{fig:batching}
\end{figure}

Figure~\ref{fig:batching} shows this pattern across batch sizes.

\subsection{Concurrency: Shared vs. Multiple Engine Instances}
\label{sec:concurrency}

Two questions followed from the batching results. Does serving concurrent requests against a single shared engine help? Does giving each request its own instance help more? Table~\ref{tab:concurrency-cpu} and Table~\ref{tab:concurrency-gpu} answer both, p50 latency and throughput for DeBERTa and Electra, shared against multi-instance, at concurrency 2, 4, and 8, each value a mean with 95 percent confidence interval across 5 independent launches.

\begin{table}[H]
\centering
\caption{CPU concurrency, p50 latency (ms) and throughput (rows/sec), mean [95\% CI], n=5.}
\label{tab:concurrency-cpu}
\resizebox{\textwidth}{!}{%
\begin{tabular}{lcccc}
\toprule
Concurrency & Shared DeBERTa & Multi DeBERTa & Shared Electra & Multi Electra \\
\midrule
2 & 646.2 [568.9, 723.4] & 676.7 [624.1, 729.3] & 394.5 [376.4, 412.6] & 352.9 [342.6, 363.1] \\
4 & 1336.9 [1302.9, 1370.9] & 2459.3 [2370.7, 2548.0] & 760.8 [733.2, 788.5] & 1800.7 [1675.7, 1925.6] \\
8 & 2632.7 [2588.9, 2676.5] & 5801.4 [5166.8, 6435.9] & 1716.6 [1648.7, 1784.5] & 3948.7 [3809.5, 4087.9] \\
\midrule
\multicolumn{5}{l}{\textit{Throughput (rows/sec), mean}} \\
2 & 1.49 & 1.35 & 2.84 & 3.09 \\
4 & 1.53 & 0.90 & 3.33 & 1.44 \\
8 & 1.65 & 0.66 & 3.05 & 1.25 \\
\bottomrule
\end{tabular}%
}
\end{table}

\begin{table}[H]
\centering
\caption{GPU concurrency, p50 latency (ms) and throughput (rows/sec), mean [95\% CI], n=5.}
\label{tab:concurrency-gpu}
\begin{tabular}{lcccc}
\toprule
Concurrency & Shared DeBERTa & Multi DeBERTa & Shared Electra & Multi Electra \\
\midrule
2 & 84.0 [79.0, 88.9] & 1522.2 [1399.9, 1644.6] & 45.1 [39.9, 50.2] & 44.7 [40.1, 49.3] \\
4 & 153.9 [149.6, 158.2] & 4339.9 [4123.7, 4556.1] & 103.8 [94.7, 112.9] & 2418.6 [2374.6, 2462.7] \\
8 & 276.8 [272.5, 281.1] & OOM, 0/5 completed & 188.0 [175.4, 200.6] & OOM, 0/5 completed \\
\midrule
\multicolumn{5}{l}{\textit{Throughput (rows/sec), mean}} \\
2 & 22.42 & 1.18 & 40.60 & 37.39 \\
4 & 24.16 & 0.84 & 35.19 & 1.68 \\
8 & 24.07 & OOM & 41.70 & OOM \\
\bottomrule
\end{tabular}
\end{table}

11 percent, at most. That's how much shared instance throughput grows from concurrency 2 to 8, across all four device and model combinations, often less, nowhere near the fourfold gain genuine parallelism at this level would predict. Latency grows roughly linearly instead, the signature of requests queueing behind one resource rather than overlapping. A third check, concurrent requests against a running FastAPI server over real HTTP, told the same story: throughput sat at 0.53 to 0.65 rows per second from concurrency 1 through 8, essentially flat. That check was a single run rather than the repeated launch methodology used elsewhere, so we treat it as confirmatory, not a headline number.

We expected separate engine instances per request to be the more realistic route to real scaling. It was worse instead, and by a wide margin. On CPU, multi-instance throughput starts close to shared's at concurrency 2 (90 to 109 percent), then declines steadily from 43 to 59 percent by concurrency 4, roughly 40 percent by concurrency 8, since each independent session spins up its own intra-op thread pool, and N sessions means N thread pools oversubscribing the same physical cores. GPU is worse still. At concurrency 2, multi-instance runs about 19 times slower than shared for DeBERTa. At concurrency 8, both models fail every one of five launches with a GPU out-of-memory error, eight independent ONNX Runtime CUDA sessions exceed this card's 6144 MiB of VRAM. A follow-up set of single, non-repeated runs at concurrency 5, 6, and 7 hit the same allocation failure at all three levels, putting the real ceiling for this model on this card closer to 4 or 5 concurrent independent sessions, not 8.

Length-aware batching is the only real lever for higher throughput on this machine's CPU and 6 GiB GPU. Not concurrency, shared or multi-instance, in either form. Whether this balance holds on hardware with more CPU cores or more GPU headroom is untested.

\begin{figure}[H]
\centering
\includegraphics[width=\linewidth]{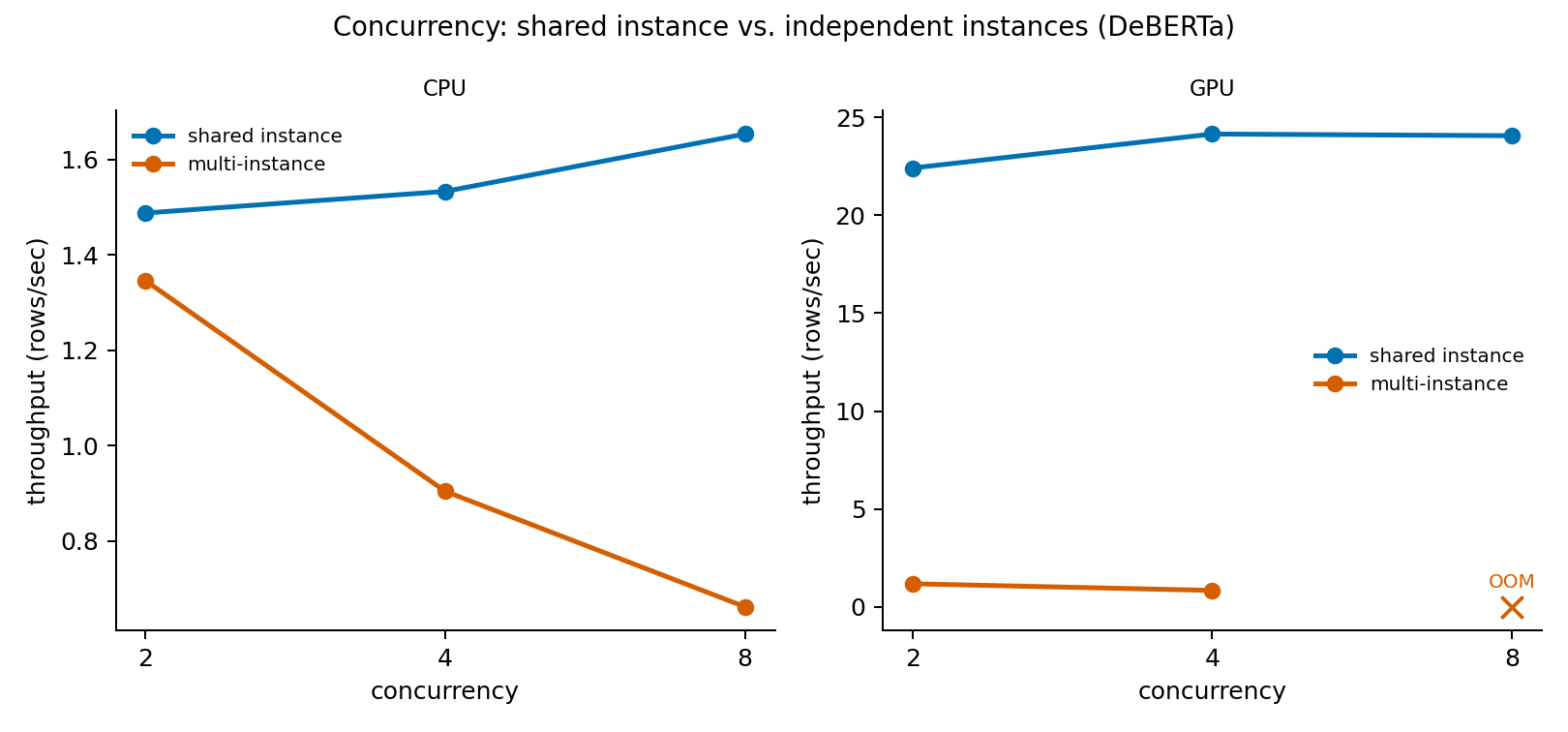}
\caption{Throughput vs. concurrency level, shared vs. multi-instance, DeBERTa.}
\label{fig:concurrency}
\end{figure}

Figure~\ref{fig:concurrency} shows the shared-instance vs. multi-instance throughput comparison, including the GPU out-of-memory failure at concurrency 8.

\subsection{Generalization: A Second Model}

WordPiece instead of SentencePiece. That's the main architectural difference in \texttt{OpenAssistant/reward-model-electra-large-discriminator}, the model we used to check whether any of the above was specific to DeBERTa. Table~\ref{tab:electra-latency} compares the two models across the three Python backends.

\begin{table}[H]
\centering
\caption{DeBERTa vs. Electra, CPU p50 latency (ms), Python backends.}
\label{tab:electra-latency}
\begin{tabular}{lcc}
\toprule
System & DeBERTa-v3 & Electra-large \\
\midrule
HF eager & 602.4$^{*}$ & 1351.0$^{*}$ \\
Python ONNX Runtime & 349.0 & 290.3 \\
torch.compile & 628.8 & 334.1 \\
\bottomrule
\end{tabular}\\
\vspace{2pt}
\footnotesize{$^{*}$HF eager row is a 3-launch measurement that did not reproduce on a later single fresh run (see discussion below); treat the DeBERTa-vs-Electra ratio in this row with caution.}
\end{table}

ONNX Runtime beats eager mode for Electra just as it does for DeBERTa, the direction generalizes cleanly. The magnitude doesn't. An earlier draft claimed the gap was substantially larger for Electra (about 4.7x) than DeBERTa (about 1.7x); we no longer stand behind that number. The obvious confound, checked first: does Electra's WordPiece tokenizer just produce longer sequences than DeBERTa's SentencePiece on the same 60 row set? No. Average token count is 221.0 for DeBERTa, 222.6 for Electra, effectively identical.

It didn't survive a fresh check either way. The HF eager numbers in Table~\ref{tab:electra-latency} came from 3 independent launches at original collection time, technically meeting this paper's repeated launch standard, but unlike every other headline comparison, this one was never independently reverified before this version. We reran it, identical script, identical 60 rows, identical machine. It didn't reproduce. DeBERTa's eager p50 came in at 725.5 ms, Electra's at 468.1 ms, Electra faster than DeBERTa, the opposite ordering from the table above. That check was a single fresh run, not a repeated launch measurement, so it can't simply replace the original numbers either. But the size and direction of the discrepancy is far larger than ordinary noise should produce on identical code and hardware.

We left the original numbers in Table~\ref{tab:electra-latency} rather than silently editing them. This is the one result in the paper that didn't get the same robustness treatment, multi-seed sampling, an independent rerun, a scaled dataset, as everything else. We don't currently trust the 4.7x versus 1.7x comparison without a fresh remeasurement we didn't have time to complete before this revision. It's a fitting demonstration of exactly the failure mode Mytkowicz et al. \cite{mytkowicz2009producing} and Georges et al. \cite{georges2007statistically} warn about, and exactly why the rest of this paper insists on repeated independent launches rather than trusting any single comparison at face value. One row slipped through that discipline. Better to flag it than leave it quietly uncorrected.

A real bug had to be fixed first to get here. Electra's initial ONNX export showed a genuine 0.003 to 0.1 absolute reward discrepancy against the PyTorch reference, traced to a missing \texttt{token\_type\_ids} input in the export wrapper. Electra has real segment embeddings, unlike DeBERTa, so every response side token was silently defaulting to segment 0. Exporting token\_type\_ids fixed it: Electra's max absolute error dropped to 2.9e-5, the same order of magnitude as DeBERTa's 5.7e-6 CPU noise floor. The latency numbers above were measured after this fix, so they're unaffected by it.

Once Electra had a native WordPiece tokenizer in the C++ engine, we reran the ablation, batching, and concurrency sweeps natively rather than through Python, same conclusions as DeBERTa across the board. The main four system comparison in Table~\ref{tab:main-latency} was not rerun with the C++ engine on Electra, though; that table stays DeBERTa only, and Table~\ref{tab:electra-latency}'s three system comparison is our generalization check for the core runtime finding specifically.

\subsection{Robustness Checks}

1.79x and 1.73x. Those are the C++-engine-to-HF-eager ratios under two different random seeds for rebuilding the dataset, with the alternate seed's HF eager latency falling squarely inside the original run's confidence interval. Neither the main latency finding nor the batching results moved outside noise on an independently sampled 150 row dataset either, about two and a half times the primary sample. Neither result looks like an artifact of one particular 60 row draw.

\section{Discussion}
\label{sec:discussion}

Engineering decisions at this layer depend entirely on the target architecture and your tolerance for compilation overhead. There is no single blanket recommendation to use C++ or to just use torch.compile.

The language wrapper is mostly irrelevant for CPU execution. Calling the identical ONNX Runtime session from Python costs the same as executing it from C++. A team scoring reward models on CPU can capture almost all of this speedup simply by exporting to ONNX and serving via Python. The custom C++ engine only contributes a narrow advantage in the form of faster tokenization. This accelerates a step that already constitutes a tiny fraction of total latency, and we observed nothing measurable from zero copy buffers or preallocation. Most of our headline CPU numbers would still hold with a much thinner implementation.

GPU execution requires a bifurcated strategy. The fact that torch.compile beats a dedicated ONNX Runtime engine at the median is backed by strict confidence intervals, not noise. For teams already embedded in the PyTorch ecosystem and willing to absorb recompilation costs on dynamic input shapes, it may simply be the superior choice. It negates the need for an export step and a standalone engine. Table \ref{tab:gpu-p95} extends this past the median to show that torch.compile wins p95 as well, by an even wider margin. This closes off the alternative theory that a static ONNX Runtime graph acts as a safer guardrail against worst case latency spikes. It does not carry that offsetting advantage here.

Recompilation risk remains the primary variable our fixed 60 row shape distribution cannot fully rule out. A rollout hitting the extreme edge of a length distribution that the compiled shape cache has not encountered will trigger a fresh compile and a massive latency spike. The repeated launches in this study never exercise that specific scenario. Whether a given rollout length distribution remains stable enough in practice for the shape cache to stay warm, and how large a cache miss spike would actually be, remains an open question best suited for a follow up study.

It is also crucial to contextualize these findings within an actual RLHF pipeline. Every metric in this section is a ratio measured on an isolated reward model forward pass. We do not directly measure what fraction of a real training step this scoring occupies relative to policy rollout generation. The closest grounded estimate sits outside this study. On a DeepSpeed Chat baseline, Xiao et al. \cite{xiao2023adaptive} report actor generation accounting for upwards of 85 percent of total RLHF step time, leaving roughly 10 percent for the training update. This is highly consistent with the motivation behind the generation focused Hybrid Engine proposed by Yao et al. \cite{yao2023deepspeedchat}. Reward scoring is not broken out explicitly in either figure. However, as a single forward pass competing against an overwhelming generation budget, it is plausibly a very minor slice of the overall wall clock pie. That estimate comes from a different model configuration and should not be read as a measurement of our specific setup. It does temper how the batching and concurrency findings below should be interpreted. They represent actionable advice for removing measured inefficiency in a specific component, not a promise that optimizing this component is where a training loop loses its wall clock time.

Defaulting to padding every request to match the longest batch member silently throttles throughput in vast swaths of existing RLHF infrastructure. It remains invisible until measured. This default behavior costs 5 to 8 times the throughput on CPU and 3.5 to 4 times on GPU compared to not batching at all. A rollout worker batching by arrival order rather than length is almost certainly leaving that throughput on the table. Sorting or bucketing by length prior to batching is computationally cheap once identified. The caveat is that this fix only pays off on GPU hardware. A CPU only deployment gets absolutely no benefit from batching.

Provisioning separate engine instances per request initially seemed like a realistic route to true scaling. It proved vastly inferior. A shared instance already serializes cleanly. Splitting into independent instances adds redundant memory and CUDA context overhead for no parallel gain, and then fails outright once VRAM runs out. On this machine CPU and 6 GiB GPU, length aware batching is the only real lever for higher throughput. Concurrency in either form is not. This balance should not be assumed to carry over unchanged to a many core server CPU or a data center GPU with far more headroom. A team designing a reward model serving layer on comparable hardware should treat adding more worker threads as a non solution and invest in batching instead.

The revelation that torch.compile beats the C++ engine, that naive batching is actively harmful rather than merely unhelpful, and that zero copy optimization disappears under repetition all point to a single conclusion. This is exactly the kind of result a single benchmark run or basic intuition gets wrong. Every number in this paper is reported as a confidence interval across independent process launches specifically to guard against that pattern.

\section{Limitations}
\label{sec:limitations}

All benchmarks were conducted on a single development machine featuring one CPU and one 6 GiB NVIDIA GPU. We lacked access to secondary environments, differing CPU microarchitectures, server class hardware, or alternative operating systems to cross reference these findings. The independently sampled 150 row dataset referenced in Table~\ref{tab:batching-cpu} serves as the closest available substitute. While it successfully rules out the possibility that our results are merely artifacts of a specific 60 row draw or a small sample size, it cannot guarantee that the CPU versus GPU dynamics or the torch.compile outcomes will hold on different silicon.

The GPU memory ceiling for multiple instance concurrency on this card hovers between 4 and 5 independent sessions. We pinned this down with single, non repeated runs at concurrency levels 5, 6, and 7 rather than the 5 launch confidence interval methodology used elsewhere. The overall direction is solid, as all three levels failed with identical out of memory errors well below the tested concurrency of 8. However, the precise boundary acts as a quick check rather than a rigorous result. It is subject to shift based on the GPU driver version, memory fragmentation from other background processes, or switching to a different card entirely.

We tested multiple engine instances exclusively as separate threads within a single process sharing one address space. Separate operating system processes would give each instance true memory isolation rather than just its own ONNX Runtime session, but this configuration was not tested. This omission should not change the underlying GPU VRAM or CPU core contention findings, since those represent strict hardware limits rather than threading artifacts. Nevertheless, process level isolation is a genuinely different configuration and remains an open variable.

We also made three deliberate exclusions. Triton Inference Server was cut entirely. An earlier scaffold version of this codebase contained a Triton baseline that computed latency and reward from a hardcoded formula instead of running a real server, and we removed it rather than pass it off as a valid comparison. A legitimate test would require a running Triton server against the exported ONNX model, which demands Docker Desktop and massive image pulls unavailable in our environment. Furthermore, multiple GPU and distributed serving setups were excluded; everything reported here relies on a single GPU on a single machine. Finally, our model architectures are strictly limited to the two OpenAssistant reward models used throughout the study. A reward model built on a different backbone or featuring substantially larger parameter counts may not exhibit the same magnitude of effect. Even so, the underlying architectural truths, such as runtime beating eager mode and naive batching degrading throughput, should generalize perfectly well.

This paper benchmarks inference in isolation rather than an end to end RLHF training loop. We do not measure wall clock training throughput with the C++ engine substituted into a live policy training run, beyond providing the existing OpenRLHF \cite{hu2024openrlhf} integration adapter that ships with the engine. The extent to which this reward model inference speedup translates into faster wall clock training time depends entirely on the fraction of a training step that reward scoring actually occupies relative to policy rollout and update. That ratio varies heavily by setup and falls outside the scope of our measurements.

\subsection{Code and Data Availability}
The C++ engine, benchmarking harness, and analysis scripts used in this study have been open sourced. They are publicly available in our GitHub repository at \url{https://github.com/vishnup22/reward-model-benchmarks}. Furthermore, the hh rlhf dataset \cite{bai2022training} utilized for all benchmarks remains completely accessible to the public.
\section{Conclusion}
\label{sec:conclusion}

We built a native C++ inference engine for reward model scoring and asked a plain systems question: does it actually beat the PyTorch options an RLHF pipeline would reach for by default, and if so, why. On CPU the answer is unambiguous. The C++ engine beats every PyTorch baseline we tested, with no overlap in confidence intervals. But the reason is not what we expected going in. Isolating the variable showed the speedup comes from running an ONNX Runtime graph instead of PyTorch eager mode, not from C++ itself. Calling the identical ONNX Runtime session from Python costs the same as calling it from C++. A team chasing this speedup does not need to write C++, they need to stop running eager mode.

On GPU the picture is genuinely mixed, not a repeat of the CPU story. The C++ engine still beats plain PyTorch eager mode and FastAPI clearly, but torch.compile edges it out at both the median and the 95th percentile. We confirmed this result holds under the same confidence interval standard as every other number in this paper, rather than trusting a single run. We report that plainly rather than explain it away, since a systems study that only reports the results it expected would not be a very useful one.

We also want to be plain about what these ratios do not tell a reader. None of them measure what fraction of a real RLHF training step reward scoring occupies relative to policy rollout generation, and outside evidence on DeepSpeed Chat style pipelines suggests generation alone can claim the large majority of a step wall clock time \cite{yao2023deepspeedchat,xiao2023adaptive}. The findings below are, we think, sound engineering advice for the component we measured. Whether that component is where a given team training loop is actually losing time is a separate question this paper does not answer.

Two further findings matter for anyone actually serving a reward model rather than just benchmarking one. Naive batch padding is not a neutral default. It actively hurts throughput on both CPU and GPU, and length aware scheduling only recovers that loss on GPU, since the CPU backend has no batch parallelism to exploit in the first place. Furthermore, scaling throughput by adding concurrent requests against a shared engine, or by giving each concurrent request its own engine instance, does not work on this hardware. Throughput stays flat under the former and gets worse under the latter. Real throughput scaling here comes only from batching.

Every one of these results, including the ones that ran counter to what we expected, held up specifically because we treated single timed runs as untrustworthy from the start and measured everything as a mean and confidence interval across independent process launches. We think that discipline, as much as any individual number in this paper, is the thing worth carrying into how RLHF infrastructure gets benchmarked going forward.

\bibliographystyle{unsrt}  
\bibliography{references}  

\section*{Acknowledgments}
We have  utilized an AI language model (Gemini) strictly as a copyediting assistant to refine sentence structure, improve readability, and format the manuscript. All experimental design, data collection, analysis, and conclusions are entirely the original work of the authors, who take full responsibility for the contents of this paper.
\end{document}